\documentclass[conference]{IEEEtran}
\IEEEoverridecommandlockouts
\usepackage{cite}
\usepackage{amsmath,amssymb,amsfonts}
\usepackage{algorithmic}
\usepackage{graphicx}
\usepackage{textcomp}
\usepackage[usenames, dvipsnames]{color}
\usepackage{balance}

\usepackage{subfigure}

\usepackage{tabularx,ragged2e}

\def\BibTeX{{\rm B\kern-.05em{\sc i\kern-.025em b}\kern-.08em
    T\kern-.1667em\lower.7ex\hbox{E}\kern-.125emX}}

\newif\ifdraft\drafttrue

\ifdraft

\fi

 \makeatletter
 \def\ps@headings{%
 \def\@oddhead{\mbox{}\scriptsize\rightmark \hfil \thepage}%
 \def\@evenhead{\scriptsize\thepage \hfil \leftmark\mbox{}}%
 \def\@oddfoot{\scriptsize \textbf{Preprint version. Accepted at IEEE ICDL-Epirob 2018.\hfil 17-20 Sep. 2018, Tokyo, Japan.}}%
 \def\@evenfoot{\scriptsize 17-20 Sep. 2018, Tokyo, Japan\hfil Accepted at IEEE ICDL-Epirob 2018}}
 \makeatother

\begin{document}

\title{Developmental Bayesian Optimization of Black-Box with Visual Similarity-Based Transfer Learning\\
{
\thanks{This work was in part supported by the EU FEDER funding through the FUI PIKAFLEX project and in part by the French National Research Agency, l'Agence Nationale de Recherche (ANR), through the ARES labcom project under grant ANR 16-LCV2-0012-01.}
}
}

\author{\IEEEauthorblockN{1\textsuperscript{st} Maxime Petit}
\IEEEauthorblockA{\textit{LIRIS, CNRS UMR 5205} \\
\textit{Ecole Centrale de Lyon}\\
Ecully, France \\
maxime.petit@ec-lyon.fr}
\and
\IEEEauthorblockN{2\textsuperscript{nd} Amaury Depierre}
\IEEEauthorblockA{\textit{LIRIS, CNRS UMR 5205} \\
\textit{Ecole Centrale de Lyon}\\
Ecully, France \\
amaury.depierre@ec-lyon.fr}
\and
\IEEEauthorblockN{3\textsuperscript{rd} Xiaofang Wang}
\IEEEauthorblockA{\textit{LIRIS, CNRS UMR 5205} \\
\textit{Ecole Centrale de Lyon}\\
Ecully, France \\
xiaofang.wang@ec-lyon.fr}
\and
\IEEEauthorblockN{4\textsuperscript{th} Emmanuel Dellandrea}
\IEEEauthorblockA{\textit{LIRIS, CNRS UMR 5205} \\
\textit{Ecole Centrale de Lyon}\\
Ecully, France \\
emmanuel.dellandrea@ec-lyon.fr}
\and
\IEEEauthorblockN{5\textsuperscript{th} Liming Chen}
\IEEEauthorblockA{\textit{LIRIS, CNRS UMR 5205} \\
\textit{Ecole Centrale de Lyon}\\
Ecully, France \\
liming.chen@ec-lyon.fr}
}

\maketitle

\begin{abstract}
We present a developmental framework based on a long-term memory and reasoning mechanisms (Vision Similarity and Bayesian Optimisation). This architecture allows a robot to optimize autonomously hyper-parameters that need to be tuned from any action and/or vision module, treated as a black-box. The learning can take advantage of past experiences (stored in the episodic and procedural memories) in order to warm-start the exploration using a set of hyper-parameters previously optimized from objects similar to the new unknown one (stored in a semantic memory). As example, the system has been  used to optimized 9 continuous hyper-parameters of a professional software (Kamido) both in simulation and with a real robot (industrial robotic arm Fanuc) with a total of 13 different objects. The robot is able to find a good object-specific optimization in 68 (simulation) or 40 (real) trials. In simulation, we demonstrate the benefit of the transfer learning based on visual similarity, as opposed to an amnesic learning (\textit{i.e.} learning from scratch all the time). Moreover, with the real robot, we show that the method consistently outperforms the manual optimization from an expert with less than 2 hours of training time to achieve more than 88\% of success.\\ \vspace{-0.1cm}
\end{abstract}


\vspace{-0.1cm}
\section{Introduction}

\begin{figure*}[ht!]
\centerline{\includegraphics[width=0.9\linewidth]{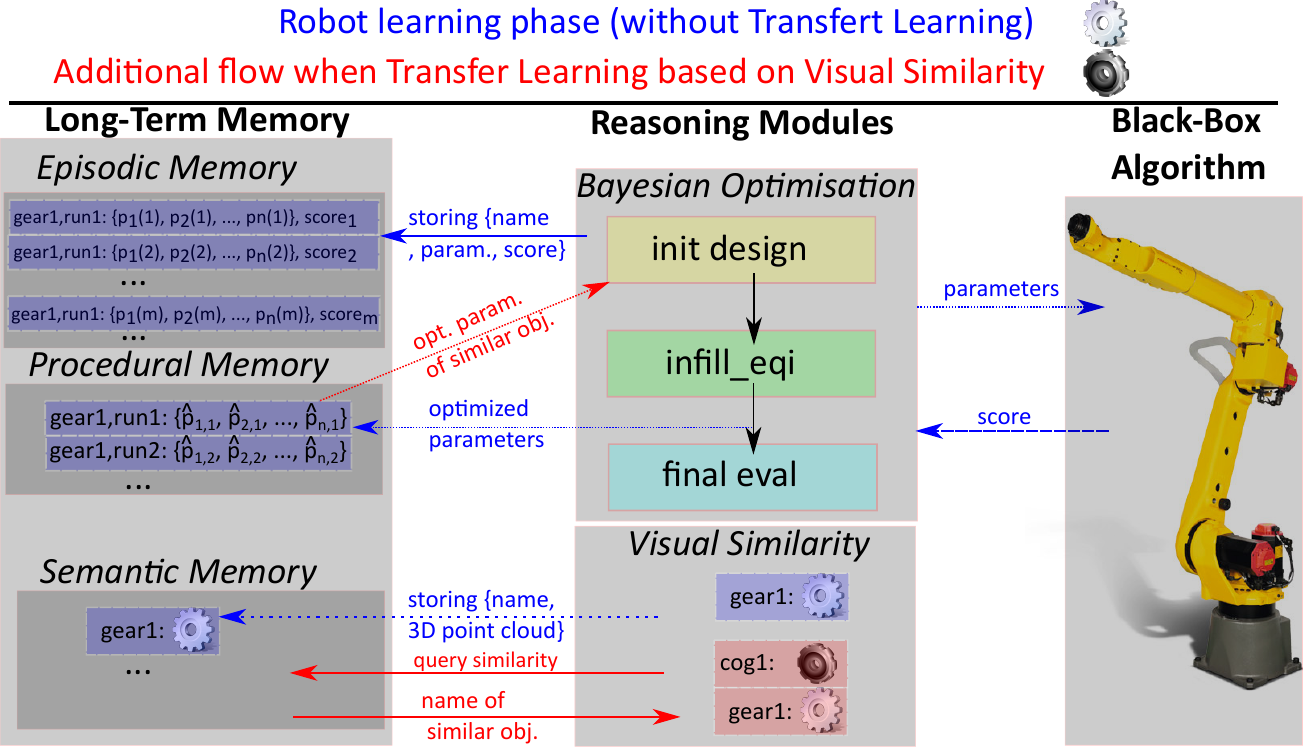}}
\vspace{-0.3cm}
\caption{Architecture of the cognitive developmental framework, based on Long-Term Memory (with episodic, procedural and semantic memories) and Reasoning Modules (Bayesian Optimisation and Visual Similarity) allowing a robot to learn how to grasp objects. This learning consists of guiding an efficient hyper-parameters optimization of black-box algorithm controlling the robot. The blue arrows represents the data flows during a learning phase without transfer learning (\textit{i.e.} without taking advantage of the Long-Term memory, just storing the experiences). The red arrows shows the additional queries and exchanges of information during a learning phase with transfer learning, based on the visual similarity between objects the robot knows how to grasp, and a new one.}
\label{fig-architecture}
\vspace{-0.5cm}
\end{figure*}





Many algorithms and frameworks in the field of robotics require optimal parameter settings to yield strong performance (\textit{e.g.} Deep Neural Networks~\cite{snoek2012practical}, Reinforcement Learning~\cite{ruckstiess2010exploring}), whether they are used to move from one place to another or to grasp objects. These parameters can be manually optimized by a human expert, but this task is tedious and error-prone. Moreover, this solution is not viable in practice for situations where the hyper-parameters have to be defined frequently, such as for each subset or task (\textit{e.g.} for each object to be grasped). To overcome these challenges, techniques have been developed to automatized high-level parameters search such as Bayesian Optimization~\cite{mockus1994}, especially suited where the evaluation of the algorithm, treated as a blackbox function, is very expensive and noisy (which is the case for real world robotic grasping application). These automated configuration techniques are however commonly used before the deployment of the solution on a system, or launched manually when needed, separated from the autonomous "life circle" (\textit{i.e.} offline) of the robotics platform from instance. Hence, the optimizations are starting from scratch each time (this is called a \textit{cold-start}) without taking advantage of the previous experience of the system (\textit{warm-start})\cite{yogatama2014efficient}, as opposed to developmental framework that mights benefit from transfer learning.

Our contribution consist of a developmental cognitive architecture (composed of a long term memory and reasoning modules) allowing a robot to optimize by experience the parameters of a manipulation and/or vision algorithm (treated as black-box) where fine-tuning according to objects is needed. The learning procedure efficiency is increased by taking advantage of previous experiences (\textit{i.e.} past optimization of similar objects). The framework will be tested in both simulation and with real robot.

\vspace{-0.1cm}
\section{Related Work}

\textbf{Bayesian Optimization} (BO)~\cite{mockus1989bayesian,brochu2010tutorial} has been used in the robotic field, especially for automatic gait optimization (\textit{e.g.} \cite{lizotte2007automatic,calandra2016bayesian,yang2018learning}). Among others, Cully \textit{et al.} developed a walking robot that can quickly adapt its  gait after been damaged using an intelligent trial and error algorithm~\cite{cully2015robots}. The robot is taking advantage of previous simulated experiences where legs were damaged, with the best walking strategies stored in a 6-dimensional behavioural space (discretized at five values for each dimension, representing the portion of time each leg is in contact with the ground). For our application, the behavioral space will have to represent the object similarity to optimize the search. However, such behavioural space cannot be objectively and numerically obtained because the concept of similarity is not easily interpretable into discrete dimensions. Regarding robot grasping tasks, the recent work of Nogueira \textit{et al.} uses BO allowing a humanoid robot iCub to not only learn how to perform grasping, but also doing it safely (\textit{i.e.} not sensitive to small errors during grasp execution)\cite{nogueira2016unscented}. However, this method only concerns single object grasping (\textit{i.e.} isolated objects) where they can be grasped in a collision-free environment from every direction, which is not the case for our context.\\
\textbf{Transfer Learning} is commonly used to reduce the time needed for a real robot to acquire new skills, usually involving reinforcement learning domain (see Taylor \textit{et al.} for a review~~\cite{taylor2009transfer}). In particular, the most recent work is from Breyer \textit{et al.} with a sim-to-real transfer based reinforcement learning for a grasping robot~\cite{breyer2018flexible} which also has the restriction that the objects needs to be isolated. The idea of transfer learning has also been applied for Bayesian Optimisation techniques: Feurer \textit{et. al.}~\cite{feurer2015initializing} used meta-learning consisting of speeding up the BO by starting from promising optimisation results that performed well on similar datasets. We are following here the same strategies but applied on real robotics using the similarity between objects.\\
Among the work on \textbf{Long-Term Memory} for robots (see review from Wood \textit{et al.}~\cite{wood2012review}), some studies were using the memory to improve the learning. Recently, a lifelong autobiographical memory has been proposed for the humanoid robot iCub~\cite{petit2016} allowing reasoning modules to stores and collect multi-modal data. Initally focusing on the declarative (episodic and semantic) memory, the framework has been extended with a procedural memory~\cite{petit2016procedural} (where skills are stored) of pointing actions.
A framework integrating declarative episodic and procedural memory systems for combining experiential knowledge with skillful know-how has also been proposed in \cite{vernon2015prospection}, based on joint perceptuo-motor representations. However, the procedural memory consist in a simple repository of pre-defined elementary actions (reach, push, grasp, locomote and wait) instead of growing flexible and adaptable skills.






\begin{figure*}[ht!] 
\vspace{-0.2cm}
\centerline{\includegraphics[width=0.90\linewidth]{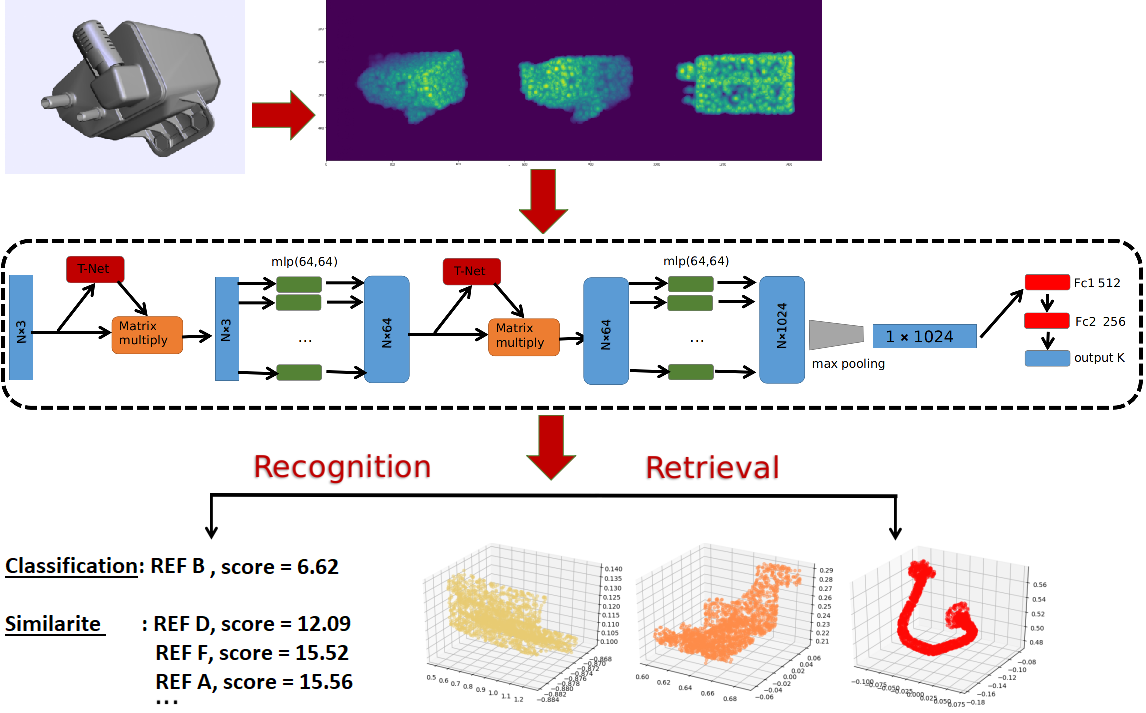}}
\vspace{-0.3cm}
\caption{Main architecture of  the Visual Similarity module. A 3D CAD model is first sampled randomly and normalized into set of points $(x,y,z)$, then it is fed into a deep neural network based on PointNet, which learns a global geometry shape by aggregating results of all points into a 1024 feature vector.      }
\label{fig-wide}
\vspace{-0.5cm}
\end{figure*}

\vspace{-0.1cm}
\section{Methodology}

\vspace{-0.1cm}
\subsection{Architecture Overview}

The architecture of the cognitive system (see Fig.~\ref{fig-architecture}) relies on an Long-Term Memory that stores information in 3 different sub-memories (as described by Tulving~\cite{tulving1985memory}): 
1) personally experiences events, that can be localized precisely in time, in the \textit{episodic memory}, 2) motor skills or action strategies in the \textit{procedural memory} and 3)  facts and knowledge about concepts and objects in the \textit{semantic memory}.
The information stored in such memories is built and accessed  by reasoning modules (a Bayesian Optimisation, see Sec.\ref{BO}, and a Visual Similarity component, see Sec.\ref{VS}). This allows the robot to take advantage of the growing knowledge acquires from experiences.
Indeed, We consider as a single experience the performance score given by the robot (\textit{i.e.} a percentage of successful grasps) using a specific set of hyper-parameters that was chosen by the Bayesian Optimisation module to be explored. The \textit{episodic memory} is thus composed by tuples with the label of the object, the set of hyper-parameters used and the obtained performance. The best set of hyper-parameters at the end of a run (\textit{i.e.} an full session of an Bayesian Optimisation) are stored in the \textit{procedural memory}, along their performance and the corresponding object. Each set represent a strategy for analysing the scene and grasping the object that has been defined by and is thus adapted to the robot used. The \textit{semantic memory} focuses here on the visual component where the 3D points clouds of objects are stored with their names by the Visual Similarity module.
This module is also able to provide the more similar objects known by the robot against another one. The Bayesian Optimisation module queries and push data from and to the episodic and procedural memory of the robot. Indeed, the transfer learning will be employed for manipulation tasks on objects (\textit{e.g.} grasping) where similar domains for the transfer implies similar objects to be manipulated. The combination of both these reasoning modules gives to the robot the capability of Transfer Learning: when confronted to a new object, the robot will first use the Visual Similarity component to extract the labels of known objects with similar shape. It is then able to access the sets of optimized hyper-parameters for these objects in order to force the Bayesian Optimisation module to explore such strategy during the initial design.  

\vspace{-0.1cm}
\subsection{Bayesian Optimisation module}\label{BO}

For the BO module, we are using the R package \textit{mlrMBO}\footnote{https://github.com/berndbischl/mlrMBO}~\cite{mlrMBO} using Gaussian Processes (GP), also known as the Kriging model~\cite{kriging} as surrogate model. It is the most often used surrogate model to estimate both the fitness (used to exploit) and the uncertainty (used to explore)\cite{j2006design}. A BO run will provide a single optimized set of parameters and can be decomposed in 3 main parts:
\vspace{-0.1cm}
\begin{itemize}
	\item initial design (\textit{"init\_design"}): select points independently to draw a first estimation of the objective function. 
    \item Bayesian search mechanism (\textit{"infill\_eqi"}), balancing exploitation and exploration, where the next point is extracted from the acquisition function (constructed from the posterior distribution over the objective function) with the Expected Quantile Improvement (EQI) criteria from Pichney \textit{et al.}\cite{Picheny2013}. It is an extension of the Expected Improvement (EI) criteria for heterogeneously noisy functions, where the improvement is measured in the model rather than on the noisy data. 
    \item final evaluation (\textit{"final\_eval"}), where the best predicted set of hyperparameters (prediction of the surrogate, which reflects the mean and is less affected by the noise) is used several times in order to provide a distribution of the optimization results.
\end{itemize}

It has to be noted that the BO is dealing with $n$ hyper-parameters $p_1, p_2, ..., p_n \in [0:1]$, which are transformed using the boundaries of the hyper-parameters before sending them to the robot. The BO module then launches the robot in order to achieve a task (\textit{e.g.} try to grasp X times an object) which provides, in return, a performance score $s_i \in [0:100]$ (\textit{i.e.} the percentage of successful grasps among K attempts), where $i$ is the iteration number. 

\vspace{-0.1cm}
\subsection{Visual Similarity module}\label{VS}

The Visual Similarity module (see Fig.~\ref{fig-wide}) is retrieving the most similar objects from semantic memory, \textit{i.e.} a retrieval and classification system learned from a reference database, where each reference's parameters has been optimized. To train such semantic memory, we apply a deep learning neural network, such as PointNet~\cite{pointnet} on 3D model. PointNet is a deep Learning method for 3D Classification and Segmentation, designed to learn point clouds geometrical shape in 3D. It takes the coordinates of $N$ points as input, that are transformed with an affine transformation matrix by a mini-network called T-Net. Then, each point is learned by convolutional kernel (size of 1), and finally aggregated by symmetric operations (\textit{i.e.} max-pooling), into a global feature of dimension 1024, followed that, three fully connected layers are stacked on to learn the  object classes.

To learn the semantic memory, our reference 3D CAD models are represented first with point clouds. Then the last layer in PointNet is modified into our reference database (change number of classes), and we  further fine-tune the PointNet by freezing earlier layers of \emph{conv5}, based on pre-trained model from ModelNet40 \cite{wu20153d}, which turns out a satisfactory accuracy and efficiency.  

To retrieve the most similar models from a new given 3D model, we first sample the new object into point clouds ($N$ points of $(x,y, z)$), then we extract its global feature with 1024 dimensions, and finally we calculate its pair of distance between each reference model, with the most similar reference corresponding to the minimal distance.  


\vspace{-0.1cm}
\subsection{Memory}

The memories are stored in a PostgreSQL\footnote{http://www.postgresql.org/} database similarly to other implementation of long-term memory system~\cite{pointeau2014,petit2016}. The episodic memory stores for each iteration $i$ of each run $r$ the label of the object, the set of $n$ hyper-parameters $\{p_1(i), p_2(i), ..., p_n(i)\}$ tested and the score $s_i$ of the performance. The procedural memory is only composed by the best set of hyper-parameters for each run $r$, $\{\hat{p}_{1,r}, \hat{p}_{2,r}, ..., \hat{p}_{n,r}\}$, along the object name learned in the run. The semantic memory contains the visual information (stored as points clouds) about the objects in order to recognize them.


\vspace{-0.1cm}
\subsection{Transfer Learning}

The transfer learning process is implementing a human-like strategy, as proposed by Feurer \textit{et. al.}~\cite{feurer2015initializing} and called meta-learning,  which consists of warm-starting the BO by recovering previously optimized set of hyper-parameters for similar tasks (\textit{i.e.} similar datasets for~\cite{feurer2015initializing}) and try them before the bayesian search mechanism. While similarity was defined by the human in their work, we will use our Visual Similarity module to provide the most similar known object (compared to a new one) which allows the robot to query its procedural memory for such object, and X different sets of optimized hyper-parameters (obtained from X different runs) are transmitted to the BO module. They will be tested at the end of the the "init$\_$design", where the other previous iterations are still proposed by the maximimLHS function. It has to be noted that in order to estimate the effect of the transfer learning, the number of total points during the "init$\_$design" will be fixed.

\vspace{-0.1cm}
\section{Experiments}

\begin{figure}[ht!]
\centerline{\includegraphics[width=0.95\linewidth]{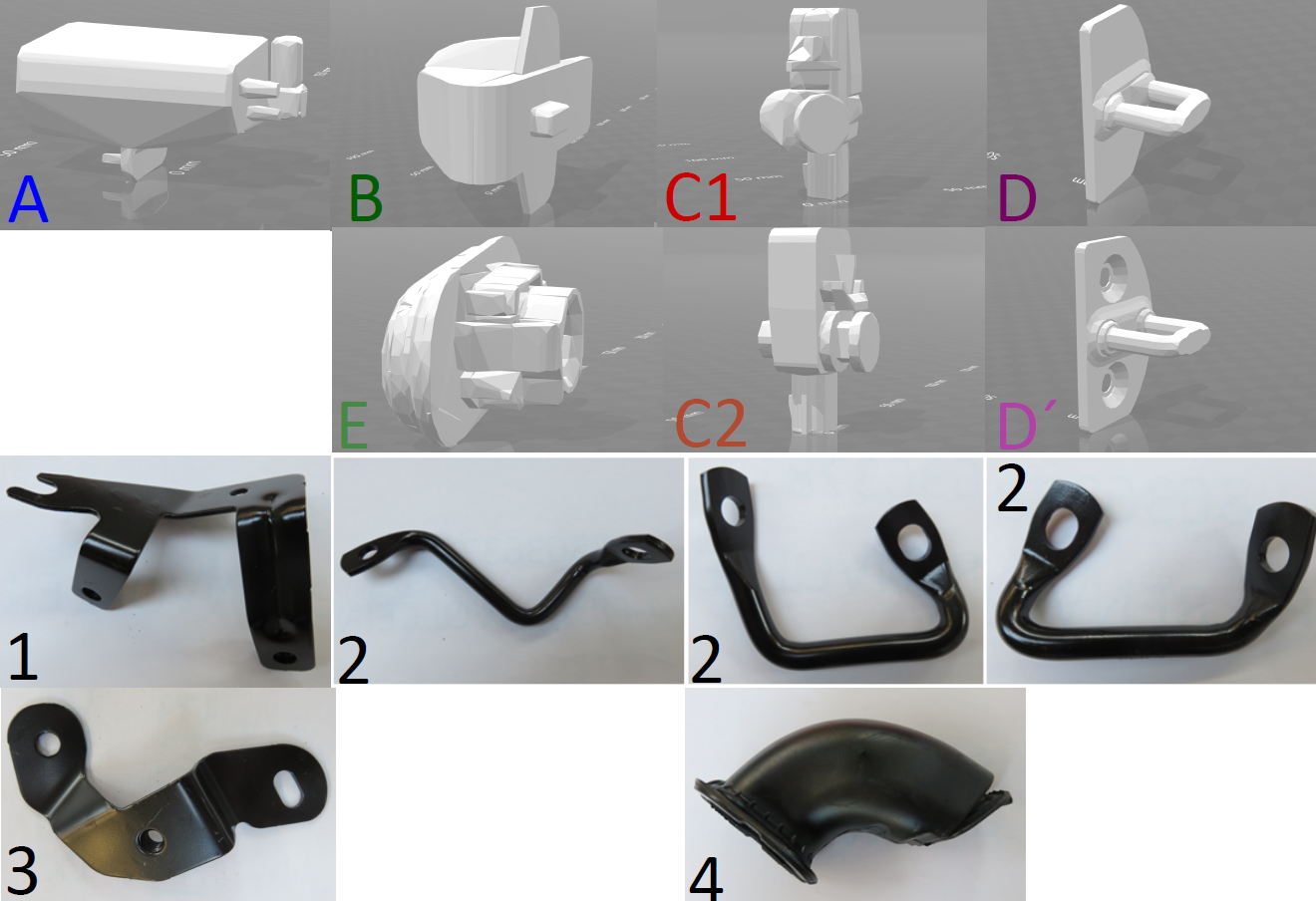}}
\vspace{-0.3cm}
\caption{Reduced CAD models of the objects used in simulation (top) and pictures of the objects used in experiments with a real robot (bottom).}
\label{fig-objects}
\vspace{-0.4cm}
\end{figure}

The task is to grasp 15 times from an homogeneous cluttered bulk composed of the same object instance (simulated or real, see Fig.~\ref{fig-objects}). We are using a professional software called Kamido\footnote{http://www.sileane.com/en/solution/gamme-kamido} from the company Sileane as a black-box controlling a parallel-jaw gripper robot that need to get parametrized for each object to manipulate. 
We will confront the framework with the optimization of 9 continuous hyper-parameters, used by Kamido to 1) analyse the images from the scene and 2) define a proper grasping target.

For the experiment in simulation, we authorize a finite budget (to be able to compare each run with the same learning conditions) of 80 iterations for the BO process, with a decomposition in 18/50/12 iterations for the 3 steps (9 parameters). The \textit{init\_design} points are selecting using a Maximin Latin Hypercube function~\cite{lhsMaximin} to maximize the minimum distance between points in order to cover space as much as possible (initially forcing the exploration). The GP's kernel is a Matern 3/2 classically used in machine learning (as explained in ~\cite{matern32}, p85). The EQI criterion is set with a quantile level $\beta = 0.7$. It has been experimentally obtained, and allows the algorithm to increase the optimization at the \textit{infill\_eqi} step after reaching a limit during the initial design step, as shown in Fig.~\ref{fig-curvePropType}, when considering 5 different objects and optimized multiple times (between 3 to 12 runs). To optimize the infill criterion, we are using a Covariance Matrix Adapting Evolutionary Strategy (CMA-ES)~\cite{cmaes1,cmaes2} from the package \textit{cmaes}\footnote{https://cran.r-project.org/package=cmaes}. It is a stochastic derivative-free numerical optimization algorithm for difficult (non-convex, ill-conditioned, multi-modal, rugged, noisy) optimization problems in continuous search spaces. 

\vspace{-0.1cm}
\section{Results}

\vspace{-0.1cm}
\subsection{From simulated Robot}

The simulator is based on the real-time physics pyBullet. Objects are loaded in the environment from their CAD model, on which we apply a volumetric hierarchical approximate convex decomposition \cite{vhacd16} in order to reduce the complexity of the collisions.

Simulation results will be shown using curves and boxplots. For each run we calculate the third quartile (Q3) of the already explored points at each iteration and compute their maximum up to now for each run (instead of the maximum directly because the evaluation is very noisy), before eventually calculating the mean among the repetitions (\textit{i.e.} same objects or With/Without transfer learning depending on the figures). The curves corresponds to a smoothing among the obtained points, using the non-parametric LOcally weighted RegrESSion (\textit{i.e.} "loess") method\cite{loess}. We also reinitialized at the beginning of each new step (\textit{i.e.} initial design, Bayesian optimisation using EQI criteria, final evaluation) to have independent points (\textit{e.g.} \textit{final\_eval} curves are not influence by the previous step).The boxplot are representing the mean of all the final performances of the optimized parameters for each run, grouped by their corresponding object.\\

\vspace{-0.8cm}
\begin{figure}[hb!]
\centerline{\includegraphics[width=0.90\linewidth]{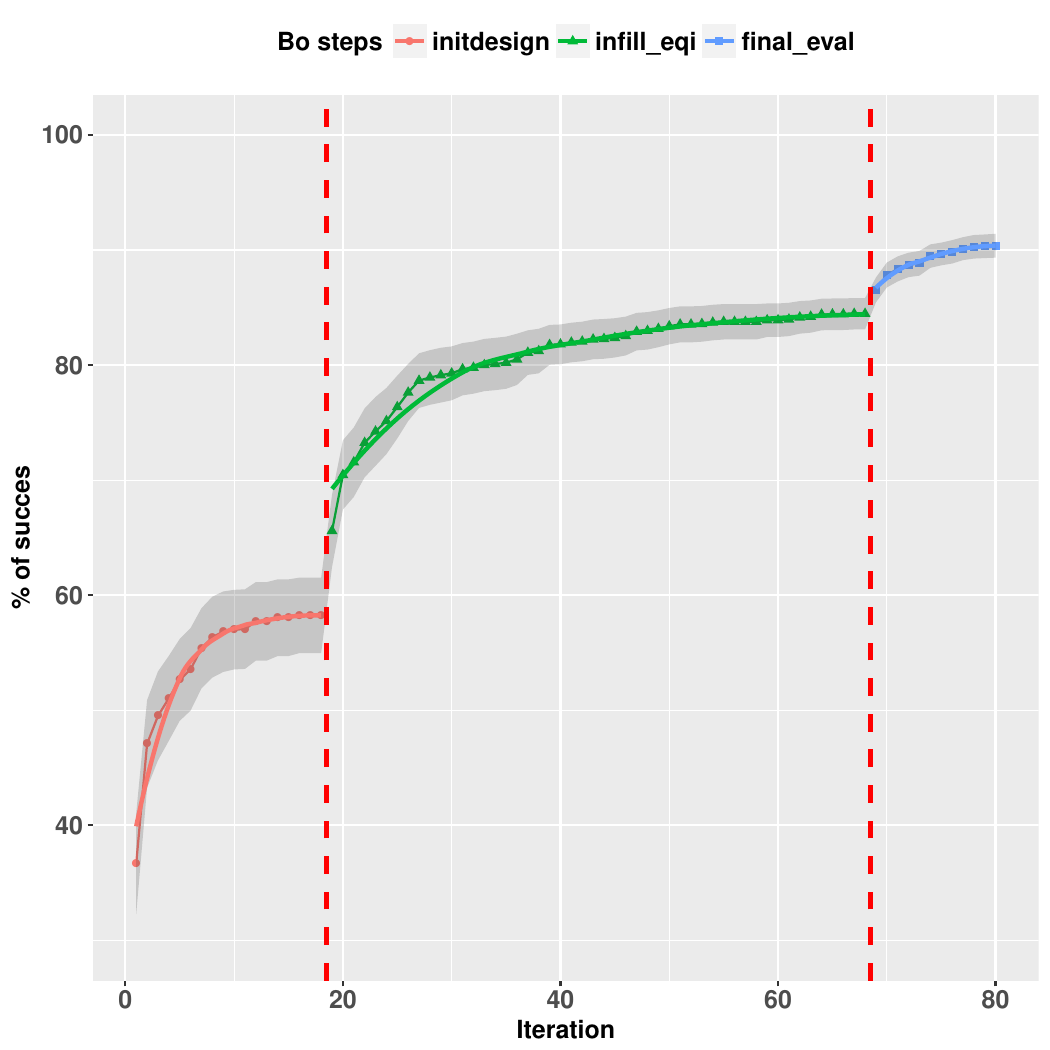}}
\vspace{-0.3cm}
\caption{Mean of the max(Q3) of the \% of success for all objects and runs combined (without transfer learning), and for each BO steps, with EQI's $\beta=0.7$}
\label{fig-curvePropType}
\vspace{-0.3cm}
\end{figure}

\subsubsection{Experiment 1, without Transfer Learning}

We show here how well the robot can develop its procedural memory to manipulate objects by optimizing its grasping hyper-parameters without using any prior information (\textit{i.e.} a "cold-start", without transfer learning)
We have tested 5 different objects in simulation: A, B, C1, C2 (\textit{i.e.} very similar to C1) and D (see Fig.~\ref{fig-objects}). We have several repetitions for each objects, in order to check the robustness of the system, respectively 6, 6, 12, 3 and 12.

Fig.~\ref{fig-curvePropType} shows an overview of the BO performance during the iterations, split among the 3 parts (\textit{i.e.} \textit{init\_design}, \textit{infill\_eqi}, \textit{final\_eval}), with objects and runs without transfer learning combined. It demonstrates that the robot benefit from the BO \textit{infill\_eqi} part with an increase of the performance thanks to the smart exploration and exploitation trade-off of the method.

\begin{figure*}[!ht]
\centering
\subfigure[Mean among the runs for each object of the Max(Q3) of the \% of success, when considering the \textit{infill\_eqi} step.]{\label{fig-simBoRes-infill}\includegraphics[width=0.425\linewidth]{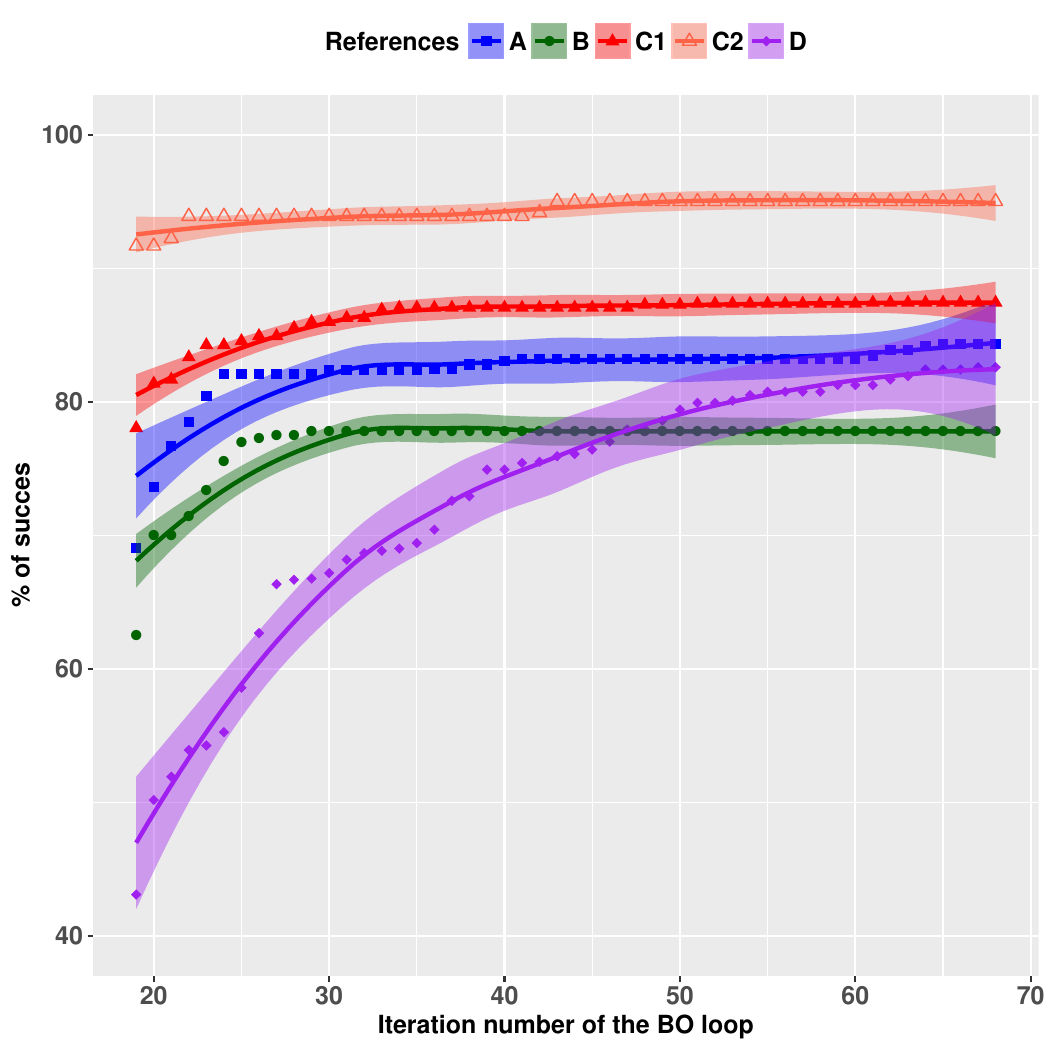}}
\hspace{0.4cm}
\subfigure[Boxplot showing the \%of success of the optimized parameters, during the \textit{final\_eval} step.]{\label{fig-simBoRes-boxplot}\includegraphics[width=0.425\linewidth]{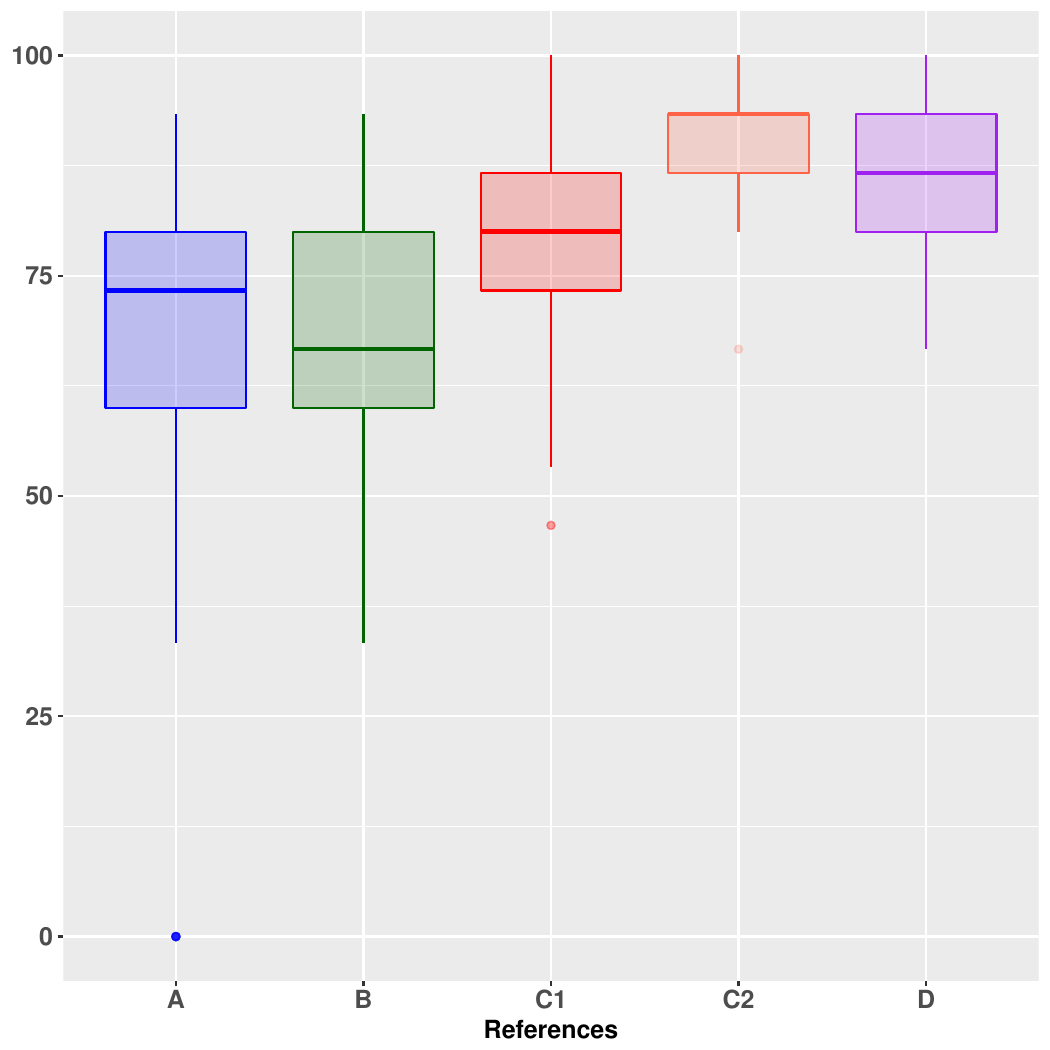}}
\vspace{-0.3cm}
\caption{Detailed results of the cold-start experiences from simulation, with an overview of the different performances for each object. Best viewed in colors.}
\label{fig-simBoRes}
\vspace{-0.5cm}
\end{figure*}

Fig.~\ref{fig-simBoRes} provides a detailed look at the results, by 1) focusing on the \textit{infill\_eqi }process and on the final evaluations, and 2) by splitting the results among the different object. In Fig.~\ref{fig-simBoRes-infill} we show the consistent increasing of performance during the 50 iterations of the BO \textit{infill\_eqi} part. The object D is the hardest to optimize and need the full 50 iterations to reach a maximum, where for the other objects the method could find one in roughly 20 iterations only.
Fig.~\ref{fig-simBoRes-boxplot} shows the final results of the obtained optimized sets of parameters for each run, grouped by objects, and completed by numerical results from Table~\ref{tab-resSim}. Object B is more difficult than the others but still achieve a decent optimization with a median score of 66.67\% (all runs combined) and 73.34\% (best run). The optimization of the objects A, C1, C2 and D are better, with median score of respectively 73.33\%, 80\%, 93.33\% and 86.67\% (all runs) and 83.33\%, 86.67\%, 93.33\% and 90\% (best run).

\subsubsection{Experiment 2, with Transfer Learning}

In these experiments, we demonstrate the effectiveness of the Visual Similarity component allowing the robot to take advantage of its acquired procedural memory to optimize faster and better with a Transfer Learning strategy when confronted to new objects. We define the number of configuration coming from such method to 3, with the parameter sets proposed by the maximinLHS method reduced to 15 in order to keep an "init$\_$design" phase with the same number of explored strategies.

\begin{figure}[htbp]
\centerline{\includegraphics[width=0.90\linewidth]{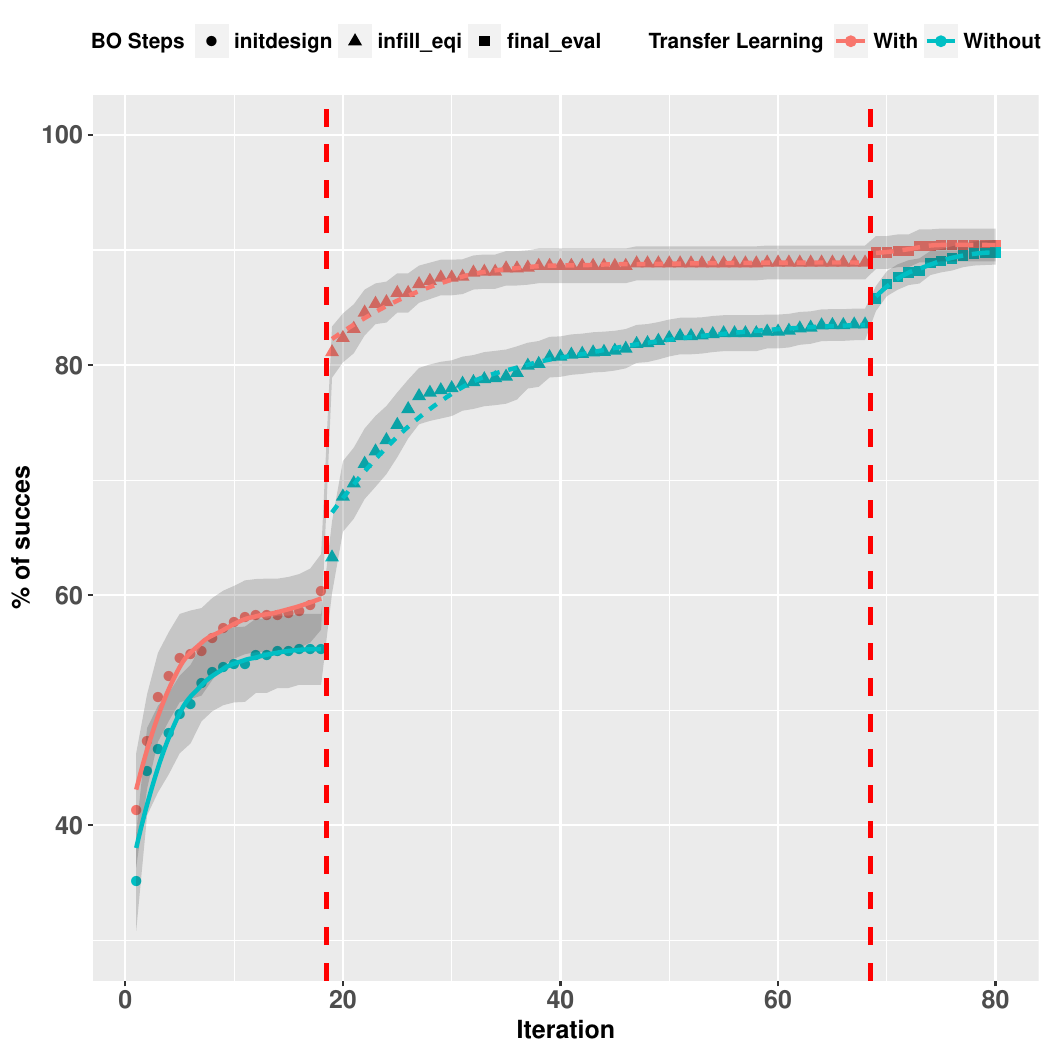}}
\vspace{-0.3cm}
\caption{Comparison With/Without Transfer Learning: Mean of the max(Q3) of the \% of success for all objects \& runs combined, and for each BO steps. Best viewed in colors.}
\label{fig-curvePropType-TL}
\vspace{-0.3cm}
\end{figure}

\begin{figure*}[!ht]
\centering
\subfigure[Transfer Learning: Mean among the runs for each object of the Max(Q3) of the \% of success, during the \textit{infill\_eqi} step.]{\label{fig-simBoRes-TL-infill}\includegraphics[width=0.425\linewidth]{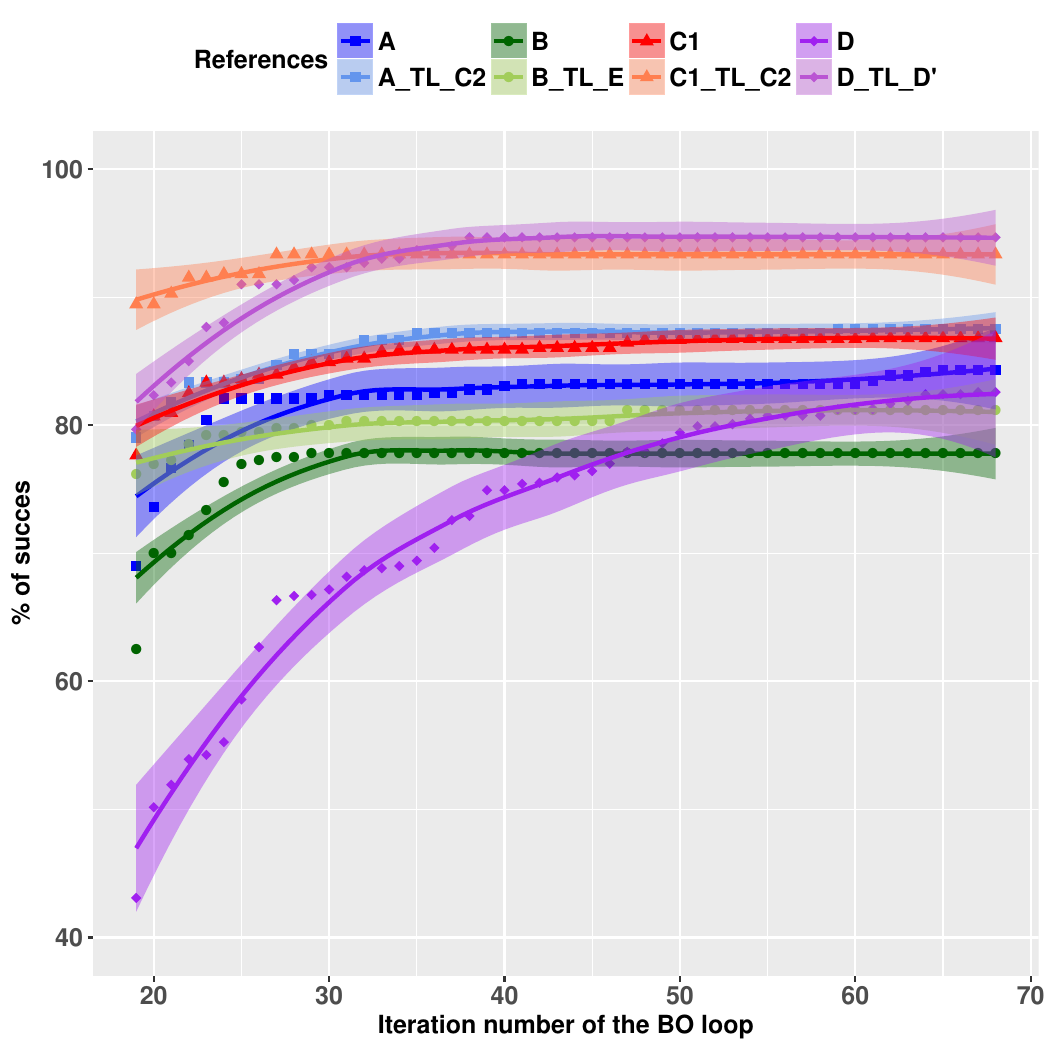}}
\hspace{0.4cm}
\subfigure[Transfer Learning: Boxplot showing the \%of success of the optimized parameters, during the \textit{final\_eval} step.]{\label{fig-simBoRes-TL-boxplot}\includegraphics[width=0.425\linewidth]{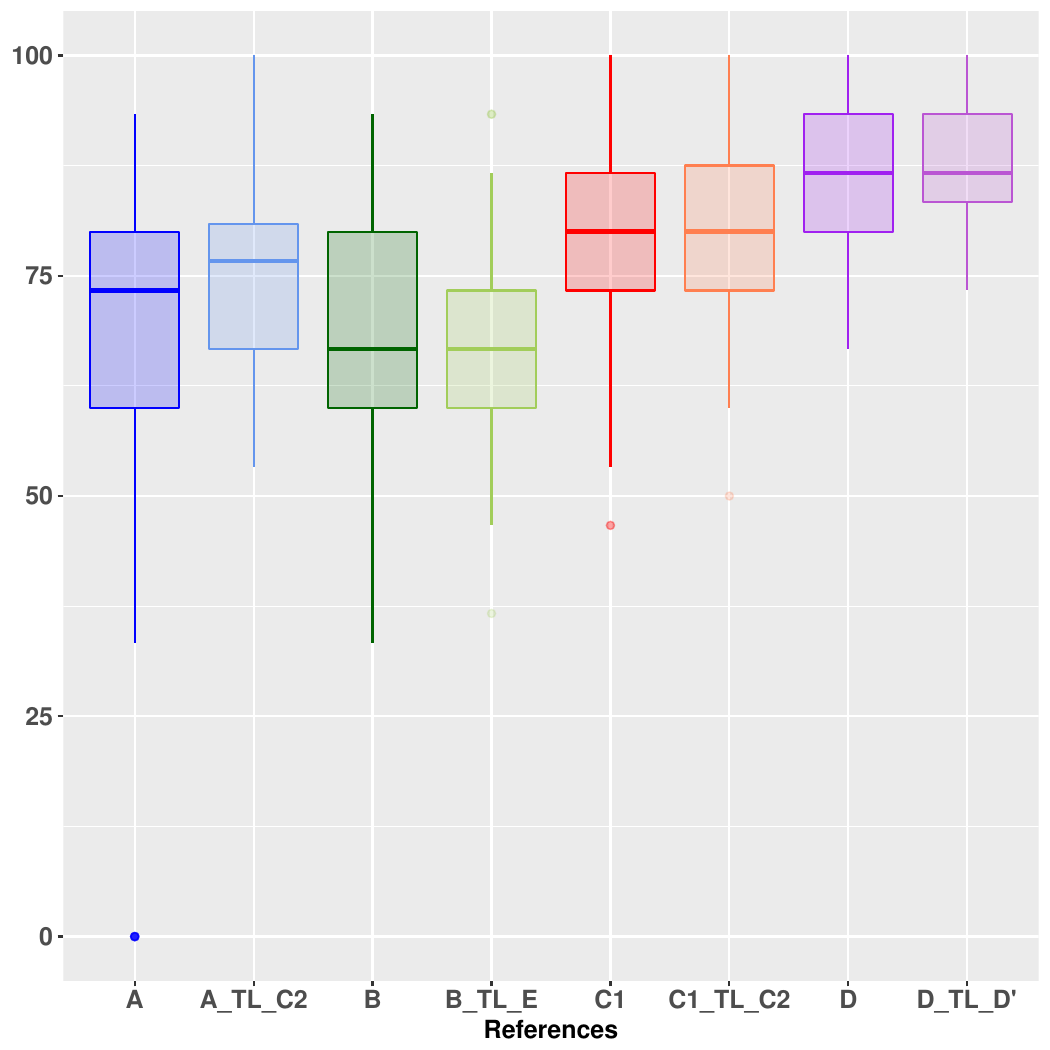}}
\caption{Detailed results of the comparison between cold-start and warm-start (\textit{i.e.} without and with transfer learning) experiences from simulation, with an overview of the different performances for each object and learning. Best viewed in colors.}
\label{fig-simBoRes-TL}
\vspace{-0.4cm}
\end{figure*}

We exemplify the concept of visual similarity and transfer learning on object D: as all other objects, the simulated model has been obtained from the volumetric hierarchical approximate convex decomposition of the original CAD model in order to reduce the complexity of collisions. For the transfer learning, we use the optimized sets of hyper-parameters obtained from experiments with the unmodified version of the object, called D', in order to shows the effect of transfer learning when using experience from very similar object. The objects A, B and C1 will be also optimized using the procedural memory of previously learned objects that are the most similar to them, respectively C2, C2 and E (corresponding scores are shown in Table~\ref{tab-simScore}). It has to be noted that whereas the pairs D/D' and C1/C2 are highly similar, the couples A/C2 and B/E are more loosely related.

Fig.~\ref{fig-curvePropType-TL} shows the performance during the BO iteration where the runs are grouped between cold-start (\textit{i.e.} without transfer learning, object A, B, C and D) and with the transfer learning based on similarity (\textit{i.e.} A\_TL\_C2, B\_TL\_E, C1\_TL\_C2 and D\_TL\_D'). At the beginning of the "infil\_eqi" part, the robot is performing better with transfer learning, and is reaching higher optima. These results are confirmed in Fig.~\ref{fig-simBoRes-TL} where the runs are grouped among their respective objects (including the differentiation between cold-start and with transfer learning, in pastel). The optimization is consistently faster to converge and with a higher optima when using transfer learning compared to without: the robot explores in a more efficient manner.

As shown in Table~\ref{tab-resSim} the optimization is better overall (\textit{i.e.} all runs for each object grouped together) with transfer learning as opposed to without for every tested object. This benefit is confirmed even if we consider only the best run per objects: when conditions are equals (6 runs for the object A and B, with/without transfer learning) the best run from transfer learning achieves a better performance than the best run from without transfer learning. On the other hand, with transfer learning, the best optimized  set of hyper-parameters achieve similar performance than the corresponding counterpart without transfer learning, despite having twice less independent runs.

\begin{table}[htbp]
\caption{Most similar object known to the robot compared to a requested object, with the corresponding similarity}
\begin{center}
\vspace{-0.3cm}
\begin{tabularx}{\columnwidth}{|c|c|X|}
\hline
Object Requested & Most Similar Object & Similarity Score\\ \hline \hline
A & C2 & 16.35 \\ \hline
B & E & 12.09 \\ \hline
C1 & C2 & 11.63 \\ \hline
\end{tabularx}
\label{tab-simScore}
\end{center}
\vspace{-0.3cm}
\end{table}

\begin{table}[htbp]
\begin{center}
\caption{Optimization Results from Simulation with Transfer Learning}
\vspace{-0.3cm}
\begin{tabularx}{\columnwidth}{|c|c|X|X|}
\hline
Ref. & Nb Runs &$\%$ success (all run, mean$\pm$sd, median)& $\%$ success (best run, mean$\pm$ sd, median)\\ \hline \hline
A & 6 &65.47$\pm$27.3, 73.33 & 78.89$\pm$11.31,83.33 \\ \hline
A\_TL\_C2 & 6 &\textbf{76.11$\pm$10.19, 76.67} & \textbf{82.78$\pm$9.93}, 83.33 \\ \hline \hline
B & 6 &68.01$\pm$12.89, 66.67 & 72.78$\pm$14.34, 73.34   \\ \hline
B\_TL\_E & 6 &\textbf{68.11$\pm$12.17}, 66.67 & \textbf{73.06$\pm$13.52, 76.67}   \\ \hline \hline
C1 & 12 &78.95$\pm$10.87, 80 & \textbf{83.89$\pm$7.63}, 86.67  \\ \hline
C1\_TL\_C2 & 6 & \textbf{81.30$\pm$11.04}, 80 & 82.5$\pm$11.82, \textbf{88.33}  \\ \hline \hline
D & 12 &86.92$\pm$9.45, 86.67 & \textbf{91.11$\pm$8.21}, 90   \\ \hline
D\_TL\_D' & 6 & \textbf{87.33$\pm$7.44}, 86.67 & 90.56$\pm7.76$, 90   \\ \hline \hline
C2 & 3 &91.76$\pm$6.74, 93.33 & 95$\pm$4.14, 93.33  \\ \hline
\end{tabularx}
\label{tab-resSim}
\end{center}
\vspace{-0.3cm}
\end{table}


\subsection{From real Robot}

In this experiment, we will compare the results of the set of hyperparameters  optimized from the Bayesian algorithm to the one manually defined by an expert after a day of tuning, without any transfer learning capabilities. We will use 4 different items (see Fig.~\ref{fig-objects}).

We are using a 6-DOF industrial robotic arm FANUC M-20iA/12L\footnote{http://www.fanuc.eu/fr/en/robots/robot-filter-page/m-20-series/m-20ia-12l} with parallel gripper. In order to keep the training time relatively short (\textit{i.e.} less than 2 hours), we are reducing the BO to a total of 45 iterations (20, 20 and 5 iterations respectively for design, Bayesian loop and final evaluation). Table~\ref{tab-resReal} shows that our optimization method outperforms the expert settings consistently on the 4 tested items, with at least 82\% of successful grasp. Moreover, item 2 is a mix of different but similar objects put together in an heterogeneous bulk, and the robot is showing robustness, by still being able to learn a common strategy to grasp them all.

\vspace{-0.3cm}
\begin{table}[htbp]
\caption{Optimization from an expert vs algorithm}
\vspace{-0.3cm}
\begin{center}
\begin{tabularx}{\columnwidth}{|c|X|X|X|X|}
\hline
Ref. Nb.& $\%$ success (expert)& $\%$ success (method)& Time per iter. (sec)& Tot. training time (min)\\ \hline \hline
1 & 80.0 & \textbf{88.0} & 133.8$\pm21.5$ & 100.4\\ \hline
2 & 88.2 & \textbf{88.6} & 124.3$\pm49.3$ & 93.2\\ \hline
3 & 74.3 & \textbf{82.2}& 160.8$\pm48.7$ & 120.6\\ \hline
4 & 86.8 & \textbf{91.7} & 132.9$\pm25.2$ & 99.6\\ \hline
\end{tabularx}
\label{tab-resReal}
\end{center}
\vspace{-0.3cm}
\end{table}


\section{Conclusion and Future Work}


We have shown how the robot can take advantage of its experience and its long-term memory to perform transfer learning when objects are similar, both in simulation and with a real robot. In simulation, with a fixed budget of 68 trials (over which the last 50 are defined by the evaluation function of a Bayesian optimization method), we are able to optimize for each 5 objects 9 continuous hyper-parameters of an industrial grasping algorithm and achieve good performance, despite the fact that the evaluation is noisy. Indeed, we achieve a mean percentage of grasping success between 73\% and 95\% in simulation (depending on the object) with a faster convergence with transfer learning. The method has also been tested in an experiment with a real robot, where the framework has been confronted to an expert. The autonomous method provides better optimization than the manually exploration from the expert, with between 82.2\% and 91.7\% of success over 4 objects (representing an increase of success between 0.4\% and 8\% compared to the expert's optimization) despite a smaller exploitation budget (20 iterations instead of 50) to keep the optimization process below 2 hours.

In this work, we have implemented a transfer learning where a similar experience (\textit{i.e.} grasping in simulation) of an object O was used for the learning of a similar object O', with the same conditions. Another type of transfer learning might be implemented in a future work: use the experience in simulation with an object O in order to warm-start the exploration of the same object but in reality, such as studied with the balancing a cart-pole ~\cite{marco2017virtual} or grasping problems~\cite{breyer2018flexible}. 
The current method is focusing on a noisy single objective optimization, where the score is solely based on the performance in terms of success. Future work might target to extend the system for optimizing multi-objective function, where for instance the speed of execution is an additional component to take into account. Pareto front of multiple criteria will then has to be considered using additional method provided by the \textit{mlrMBO} R-package~\cite{Horn2017,horn2015model}.
Another possible benefit of our framework is that the exploration and storing of different optimized set of hyper-parameters can lead to embodied symbols or concepts emergence~\cite{taniguchi2016symbol}. By providing human labels about the physical aspect of objects (\textit{e.g.} "heavy", "flat"), co-occurences can be detected between these adjectives and a sub-set of optimized parameters, in a similar way that done to discover pronouns~\cite{pointeau2014emergence} or body-parts and basic motor skills~\cite{petit2016hierarchical}. In return, this might lead to another form of transfer learning, when some hyper-parameters will be directly extracted and fixed based on description labels of new objects provided by the human.




\balance
\bibliographystyle{IEEEtran}
\bibliography{./refs.bib}

\end{document}